\documentclass{article}
\usepackage{ijcai19}

\usepackage{times}
\usepackage{soul}
\usepackage{url}
\usepackage{hyperref}
\usepackage[utf8]{inputenc}
\usepackage[small]{caption}
\usepackage{graphicx}
\usepackage{amsmath}
\usepackage{booktabs}
\usepackage{algorithm}
\usepackage{subcaption}
\usepackage[noend]{algpseudocode}
\newcommand{\etal}{\text{et al.}}

\title{RWR-GAE: Random Walk Regularization for Graph Auto Encoders}

\author{
Vaibhav\footnote{Contact Author}\and
Po-Yao Huang\and
Robert Frederking\\
\affiliations
Carnegie Mellon University\\
\emails
\{vvaibhav, poyaoh, ref\}@cs.cmu.edu}

\begin{document}

\maketitle

\begin{abstract}
Node embeddings have become an ubiquitous technique for representing graph data in a low dimensional space. Graph autoencoders, as one of the widely adapted deep models, have been proposed to learn graph embeddings in an unsupervised way by minimizing the reconstruction error for the graph data. However, its reconstruction loss ignores the distribution of the latent representation, and thus leading to inferior embeddings. To mitigate this problem, we propose a random walk based method to regularize the representations learnt by the encoder. We show that the proposed novel enhancement beats the existing state-of-the-art models by a large margin (upto 7.5\%) for node clustering task, and achieves state-of-the-art accuracy on the link prediction task for three standard datasets, \emph{cora}, \emph{citeseer} and \emph{pubmed}. Code available at \url{https://github.com/MysteryVaibhav/DW-GAE}.
\end{abstract}

\section{Introduction}

Analysis of graph data plays an important role in various data mining tasks including node classification~\cite{kipf2016semi}, link prediction~\cite{wang2017predictive}, and node clustering~\cite{wang2017mgae}. These tasks are useful for various kinds of graph data including protein-protein interaction networks, social media, and citation networks. However, it is known that working with graph data is a challenging task because of its high computational cost and low parallelizability. Further, the inapplicability of machine learning methods~\cite{cui2018survey} to such data aggravates the problem.

\begin{figure}[t]
\centering
\begin{subfigure}{.25\textwidth}
  \includegraphics[width=3.2cm, height=2.5cm]{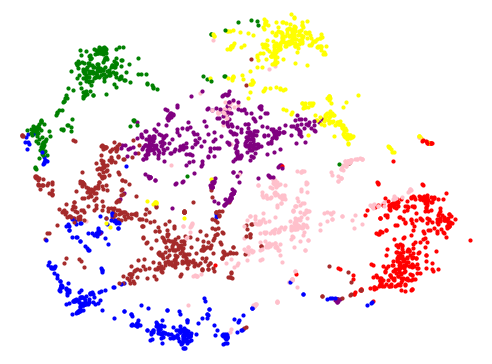}
\end{subfigure}%
\begin{subfigure}{.45\textwidth}
  \includegraphics[width=3.2cm, height=2.5cm]{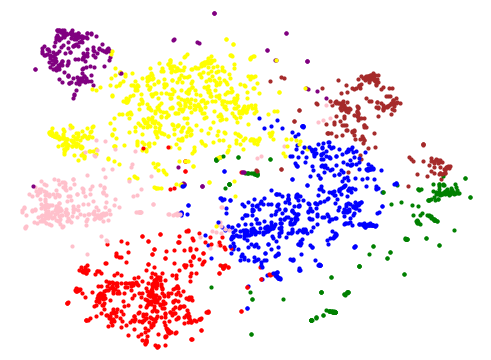}
\end{subfigure}%
\caption {Node embeddings learned by two different architectures. \textbf{Left:} Generated by an autoencoder model \textbf{Right:} Generated by the proposed model. We see that the embeddings on the left have dense representations for nodes belonging to the same cluster whereas the embeddings on the right have an even intra-cluster spread which reduces the potential loss in clustering accuracy across boundaries. See \S~\ref{graph:dis} for a detailed discussion.}
\label{fig:example}
\vspace{-1\baselineskip}
\end{figure}


Recent developments in graph embeddings have emerged as a boon for dealing with complex graph data. The general idea behind learning a graph embedding is to learn a latent, low-dimensional representation of network vertices, while preserving network topology structure, vertex content, and other information. \cite{perozzi2014deepwalk} proposed a DeepWalk model to learn node embeddings by reducing the problem to a skipgram formulation~\cite{mikolov2013distributed} used to learn word embeddings. Recent works~\cite{vgae} show that graph autoencoder in conjunction with Graph Convolutional networks~\cite{kipf2016semi} are even more effective in learning low dimensional representations of the nodes. However, there are a few shortcomings in using autoencoders for learning graph embeddings. First, there is no restriction on the distribution of the latent representation learnt by the encoder which might result in inefficient embeddings~\cite{argae}. Second, the reconstruction loss might not be a strong signal to capture the local graph topology~\cite{goyal2018capturing}. Figure \ref{fig:example} shows the effect of these problems on Cora.

\cite{argae} tried to address the first shortcoming by applying a Gaussian prior on the distribution of node representations. We argue that enforcing Gaussian prior on the latent code of node embeddings might not be the best option and propose a random walk based regularization technique which tries to enforce a restriction on the representation such that the embeddings learn to predict their context nodes. This is achieved by adding an additional training objective. This serves two purpose at once, first, instead of adding a prior on the latent representation of the nodes, we provide additional supervision for improving the quality of each node embedding. Second, the node embeddings are forced to capture the local network topology since the added objective is maximized when the node embeddings correctly predict their context embeddings. The proposed model allows for a natural graph regularization on the embeddings, whilst providing additional training signals to improve individual embeddings.

Through our experiments, we show that the proposed random walk regularization is superior to all other methods at unsupervised clustering task. The contributions of this paper are two fold,
\begin{itemize}
    \item We propose a novel technique of using random walks for regularizing the node representations learned by a Graph autoencoder.
    \item We show that the proposed regularization technique is effective at unsupervised clustering and outperforms all the other methods. Further, we show that the resulting embeddings are general in nature and achieve state of the art accuracy on the link prediction task as well.
\end{itemize}

\section{Related Work}
Learning node embeddings for networks has been a long standing problem. Conventionally, learning node embeddings was seen as either a feature engineering task or a dimentionality reduction task. \cite{tang2011leveraging} and ~\cite{henderson2011s} proposed to use hand-crafted features based on the network properties. On the other hand, \cite{belkin2002laplacian} and \cite{roweis2000nonlinear} used linear algebra tools to reduce the adjaceny matrix of a graph to a lower dimension.

The advancement of feature learning in other domains, particularly the SkipGram model~\cite{mikolov2013distributed}, proposed to learn word embeddings opened ways to learn node features as well. \cite{perozzi2014deepwalk} proposed a DeepWalk model which used random walk~\cite{rwr} for learning graph embeddings. Their proposed objective was similar to the SkipGram~\cite{mikolov2013distributed} model. They used nodes obtained from a random walk as the context nodes and tried to predict the context nodes using the node on which the walk was performed. This work exploited the graph structure to learn the embeddings. \cite{yang2015network} proposed an extension to the DeepWalk model which enhanced the node representations by additionally incorporating the node features available from other sources, like the text features for each node. Since then, a number of probabilistic models have been proposed including \cite{grover2016node2vec} and \cite{tang2015line}, which try to map the nodes to a low-dimensional space of features that maximizes the likelihood of preserving network neighborhoods of nodes.

In the current research where deep learning is taking control over everything, Graph autoencoders have emerged as the go-to method for embedding graphs, mostly because of its good performance, efficiency and ease of use. The idea of integrating graph with neural models was first introduced by \cite{kipf2016semi}, who proposed Graph Convolution Networks (GCN) which could effectively encode graphs. GCNs can naturally incorporate node features, which significantly improves predictive performance on various tasks. Inspired by the autoencoder frameworks~\cite{kingma2013auto}, Kipf~\etal~proposed Graph autoencoder framework~\cite{vgae} which used GCNs as encoder and simple inner product as decoder. \cite{argae} identified that the graph autoencoders don't put any restriction on the distribution of latent representation which could possibly lead to inferior embeddings. To address this problem, they proposed an adversarially regularized graph auto encoder which puts a Gaussian prior on the latent distribution. Our work is motivated from this work, and we argue that Gaussian prior might not be the most natural distribution for a node's latent representation. We instead propose a random walk based regularization method which doesn't enforce any prior on the latent representation but regularizes the representations in such a way that they learn the network's local topology.

\section{Problem Definition}
We consider a general problem of learning unsupervised graph embeddings for any graph $G$. A graph $G = (V, E)$ can be represented in terms of its vertices $(V = \{v_1, v_1, \dots, v_n\})$ and edges $(E=\{e_{ij} \} \forall i, j$ s.t $\exists $ an edge between the nodes $v_i$ and $v_j$. To efficiently represent the graph topology for computational use, we represent the edges using an adjacency matrix $A \in \mathcal{R}^{n \times n}$, where $A_{ij} = 1$ if $e_{ij} \in E$ else $A_{ij} = 0$. Depending on the nature of the graph, we might have an additional node feature matrix $X \in \mathcal{R}^{n \times h}$, where each row of the matrix represents a \textit{h}-dimensional content vector for each node in the graph.

Given a graph $G$, we want to learn a \textit{d}-dimension vector for each node $v_i$ such that $d << n$. Putting everything together, we want to learn a function $F$ such that $F(A, X) \longrightarrow Z$, where $Z$ is an embedding matrix in $\mathcal{R}^{n \times d}$. We want $Z$ to capture the node content as well as the topological structure in a continuous low dimensional space.

\begin{figure*}[h]
\centering
\includegraphics[width=16cm]{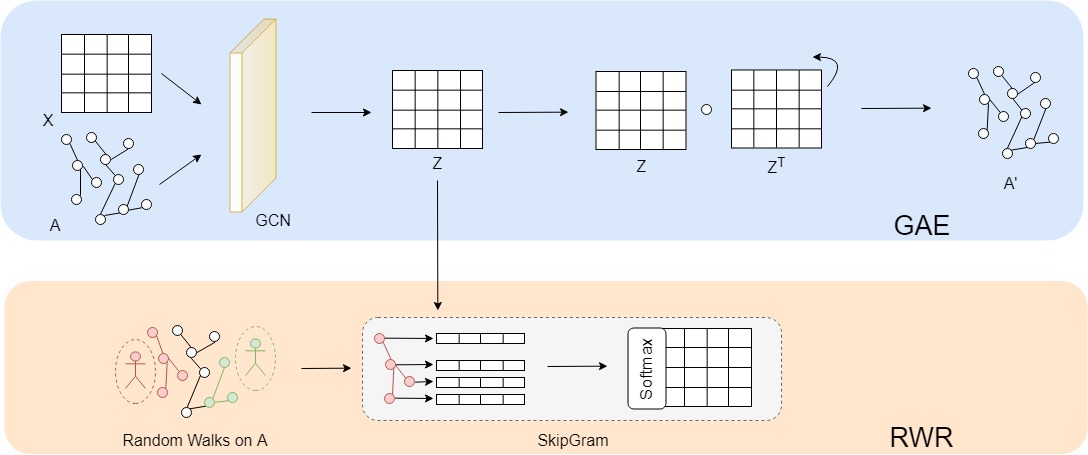}
\caption{Random Walk Regularized Graph Autoencoder. Top half of the network corresponds to the Graph Auto-Encoder. Bottom half shows the proposed Random Walk Regularization network. \label{arch}}
\end{figure*}

\section{Proposed Model}
\subsection{Graph Convolutional Networks}
\cite{kipf2016semi} introduced GCN to directly embed the graph structure in a low dimensional space using neural networks. Given a graph $\mathcal{G} = (A, X)$ where $A$ is the adjacency matrix and $X$ is the feature matrix, the graph convolutional network is a spectral convolutional operation denoted by $f(Z^l, A|W^l)$,
\begin{align}
    Z^{l+1} &= f(Z^l, A|W^l)
\end{align}
Here, $Z^l$ is the output feature corresponding to the nodes after $l^{th}$ convolution. $W^l$ is the parameter associated with the $l^{th}$ layer. Based on the above operation, we can define arbitrarily deep networks. However, one caveat to constructing deep graph convolutional networks is that, after each layer $W^l$ is multiplied with $A$ and since $A$ is not normalized, it changes the scale of the feature vectors. To address this issue, we refine the convolutional function to be,
\begin{align}
    f(Z^l, A|W^l) = \sigma(\hat{D}^{-1/2}\hat{A}\hat{D}^{-1/2}Z^lW^l)
\end{align}
Here,  $\hat{A} = A + I$, where $I$ is the identity matrix, $\hat{D}$ is the diagonal node degree matrix of $\hat{A}$ and $\sigma$ is the activation function.

\subsection{Graph Autoencoder}
Graph autoencoders are an extension to the autoencoder framework consisting of an encoder and a decoder network. We use a 2-layer GCN as the encoder and inner product as the decoder. The encoder output is given by $Z^2 = Z = q(Z|X,A)$,
\begin{align}
    Z^1 &= f_{relu}(Z^0, A|W^0)\\
    Z^2 &= f_{linear}(Z^1, A|W^1)
\end{align}
The obtained node embeddings are then used in the decoder to reconstruct the graph ($\hat{A}$),
\begin{align} \label{reconstruct}
    \hat{A} = \sigma(ZZ^T)
\end{align}
Note that we can reconstruct both $A$ and $X$. However for our method, we just reconstruct the adjacency matrix as it is more flexible for graphs which don't have content information.
The network is trained using a reconstruction loss $\mathcal{L}_R$,
\begin{align}\label{llr}
    \mathcal{L}_R = E_{q(Z|X,A)}[\log p(A|Z)]
\end{align}
\subsection{Variational Graph Autoencoder}
Variational Graph autoencoders are defined by an inference model, 
\begin{align}
    q(Z|X, A) &= \prod_{i=1}^n q(z_i|X, A)\\
    q(z_i|X, A) &= \mathcal{N}(z_i|\mu_i, diag(\sigma^2))
\end{align}
Here, $\mu = Z^2$ is a matrix of mean vectors $z_i$, $\sigma = f_{linear}(Z^1, A|W')$ is the covariance matrix. The decoder model remains roughly the same and the adjacency matrix can be reconstructed using the mean vectors, 
\begin{align}
    \hat{A}_{ij} = \sigma(z_i^Tz_j)
\end{align}
For training the variational graph autoencoder, we optimize the variational lower bound as follows, 
\begin{align}\label{lr}
    \mathcal{L}_R = E_{q(Z|X,A)}[\log p(A|Z)] + KL(q(Z|X,A) || p(Z))
\end{align}
Here, $KL(q(\cdot) || p(\cdot))$ denotes the Kullback-Leibler divergance and $p(Z) = \prod_i \mathcal{N}(z_i|0, I)$ denotes the Guassian prior for the latent data distribution. We perform the reparameterization trick~\cite{kingma2013auto} to train the variational model.

\subsection{Random Walk Regularization}
The main contribution of our model is the proposed regularization technique which forces the latent representation of the nodes to inherently capture the information of their immediate context nodes. We argue that using a Graph autoencoder with reconstruction loss for learning the node embeddings doesn't force the latent representations of the nodes to necessarily capture the local context information present at various locations in the network. Thus, we add an extra objective while training to enforce this restriction. Inspired from DeepWalk~\cite{perozzi2014deepwalk}, we leverage local information obtained from truncated random walks to learn latent representations by treating walks as the equivalent of sentences. Figure \ref{arch} shows the overall architecture of our proposed network. The lower half of the figure represents the regularization network. There are two main components of the regularization network, (a) Random Walk with Restarts and (b) SkipGram model.

\subsubsection{Random Walk with Restarts}
We leverage the idea of Random Walk with Restarts (RWR)~\cite{rwr} to obtain context nodes from any given node. $\mathcal{W}_{v_i}$ denotes a set of context nodes obtained using RWR from the start node $v_i$. Algorithm \ref{alg:rwr} defines a procedure to obtain $\mathcal{W}_{v_i}$. 

\begin{algorithm}
\caption{Random Walk with Restarts}
\label{alg:rwr}
\begin{algorithmic}
\State \textbf{Inputs: }{adjacency matrix $A$, start node $v_0$, path length $t$, restart probability $\alpha$}\\
\Procedure{Random\_Walk}{$A, v_0, l, \alpha$}
\State $ path = [v_0] $ \Comment{{\small path stores the nodes in the walk }}
\While{path length $\le$ t - 1}

\State $curr \gets$ last node in $path$
\State $ rand = Random(0, 1) $ \Comment{{ \small random number $\in (0, 1)$}}
\If{$rand \ge \alpha$}
    \State add $random\_neighbor(curr, A)$ to $path$
  \Else
    \State add $v_0$ to $path$ \Comment{{ \small in the case of restart}}
\EndIf
\EndWhile
\State \textbf{return} $path$
\EndProcedure
\end{algorithmic}
\end{algorithm}

\subsubsection{SkipGram}
Once we obtain a set of context nodes, we use a SkipGram~\cite{mikolov2013efficient} type model which has two embedding layers corresponding to the nodes and context nodes. Originally, SkipGram was designed as a language model that maximizes the co-occurrence probability among the words that appear within a window in a sentence. For this graph setting, we borrow the idea from ~\cite{perozzi2014deepwalk}, and use the set of nodes obtained from the random walk as our sentence and maximize the co-occurrence probability of the nodes. The objective function used to train this model is given by the equation below,
\begin{align} \label{sg}
    \mathcal{L}_S = \log p(\mu_i | Z(v_i))
\end{align}
Here, $\mu_i \in \mathcal{W}_{v_i}$ and $Z(v_i)$ denotes the latent representation for the node $v_i$ generated by the encoder.

\subsubsection{Our model}
Algorithm \ref{alg:algorithm} is our overall proposed framework. We train the entire network in an end-to-end fashion. An important consideration while training the network was choosing the order of the back-propagated gradients.
\begin{figure}[h]
\centering
\includegraphics[width=8cm]{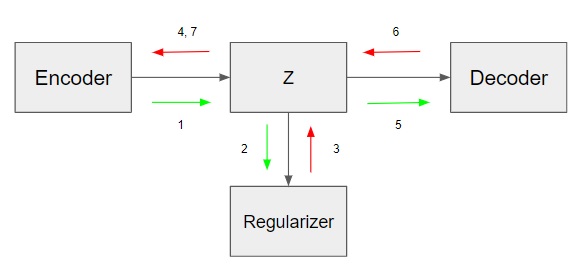}
\caption{Forward and backward propagation in order for training the model. Green arrow denotes the forward propagation and red denotes backward.\label{flow}}
\end{figure}
We experimented with all possibilities and picked the one which gave best performance as per Figure \ref{flow}. 

\begin{algorithm}
\begin{algorithmic}
\State \textbf{Input:} Graph $G(V, X, A)$, window size $w$, walks per epoch $\gamma$, walk length $t$, restart probability $\alpha$\\

\State $Z \gets$ Encoder($G$)
\State $V'$ = shuffle$(V)$
\State $\mathcal{O}$ = sample $\gamma$ vertices from $V'$
\For{each $v_i \in \mathcal{O}$}
\State $\mathcal{W}_{v_i} = $ Random\_Walk$(A, v_i, t, alpha)$
\For{each $v_j \in W_{v_i}$}
\For{each $\mu_k \in W_{v_i}[j-w : j+w]$}
\State $\mathcal{L}_{v_j} = -\log Pr( \mu_k | Z(v_j))$
\State Update SkipGram and Encoder using $\nabla \mathcal{L}_{v_j}$
\EndFor
\EndFor
\EndFor
\State $\hat{A} \gets$ Decoder($Z$)
\State Update Decoder and Encoder using Eq.\ref{llr} or Eq.\ref{lr}
\caption{Regularization through Random Walk for Graph Autoencoders}
\label{alg:algorithm}
\end{algorithmic}
\end{algorithm}

Based on the type of encoder-decoder framework used, we present two kinds of regularized network,
\begin{itemize}
    \item \textbf{Random Walk Regularized Graph Autoencoder (RWR-GAE)}, for this model we use Eq. \ref{llr} to update the decoder and encoder parameters.
    \item \textbf{Random Walk Regularized Variational Graph Eutoencoder (RWR-VGAE)}, the encoder in this model is based on the variational inference model, and we use Eq. \ref{lr} to update the decoder and encoder weights.
\end{itemize}
For both the models, we additionally use Eq. \ref{sg} for updating the skipgram model and encoder parameters.
\section{Baselines}
A rich line of work has been done for learning graph embeddings in an unsupervised setting. We briefly summarize some of the recent approaches used as our baseline,
\begin{itemize}
\item DeepWalk~\cite{perozzi2014deepwalk}: is a network representation
approach which encodes social relations into
a continuous vector space by learning structural regularities
present within short random walks.
\item Spectral Clustering~\cite{tang2011leveraging}: is an effective
approach for learning social embedding. This method generates a representation in $\mathcal{R}^d$ from the \textit{d}-smallest eigenvectors of $\mathcal{L}$, the normalized graph Laplacian of $G$.
\item GAE~\cite{vgae}: is the most recent
autoencoder-based unsupervised framework for graph
data, which naturally leverages both topological and
content information.
\item VGAE~\cite{vgae}: is a variational graph
autoencoder approach for graph embedding with both
topological and content information.
\item ARGA~\cite{argae}: is an adversarially regularized autoencoder
algorithm which uses graph autoencoder.
\item ARVGA~\cite{argae}: is also an adversarially regularized autoencoder, which uses a variational graph autoencoder.
\end{itemize}

\begin{table}[!h]
\centering{}
\begin{tabular}{c c c c}
  \toprule
 Training Data & \# Nodes & \# Edges & \# Features\\
  \midrule
	Cora & 2,708 & 8,976 & 1,433\\ 
	Citeseer & 3,327 & 7,740 & 3,703\\
	PubMed & 19,717 & 37,676 & 500\\
  \bottomrule
\end{tabular}
\caption{Statistics about different training datasets. \label{stats}}
\end{table}

\begin{table*}[t]
\begin{center}
\begin{tabular}{lcccccc}
\toprule
{\bf Model}&\multicolumn{2}{c}{\bf Cora }&\multicolumn{2}{c}{\bf Citeseer} &\multicolumn{2}{c}{\bf PubMed}\\
\midrule
& \bf AUC & \bf AP & \bf AUC & \bf AP & \bf AUC & \bf AP \\
\midrule
SC & 84.6  $\pm$  0.01 & 88.5  $\pm$  0.00 & 80.5  $\pm$  0.01 & 85.0  $\pm$  0.01 & 84.2  $\pm$  0.02 & 87.8  $\pm$  0.01\\
DW & 83.1 $\pm$  0.01 & 85.0  $\pm$  0.00 & 80.5  $\pm$  0.02 & 83.6  $\pm$  0.01 & 84.4  $\pm$  0.00 & 84.1  $\pm$  0.00\\
GAE & 91.0 $\pm$  0.02 & 92.0  $\pm$  0.03 & 89.5  $\pm$  0.04 & 89.9  $\pm$  0.05 & 96.4  $\pm$  0.00 & 96.5  $\pm$  0.00\\
VGAE & 91.4 $\pm$  0.01 & 92.6  $\pm$  0.01 & 90.8  $\pm$  0.02 & 92.0  $\pm$  0.02 & 94.4  $\pm$  0.02 & 94.7  $\pm$  0.02\\
ARGE & 92.4  $\pm$  0.003 & 93.2  $\pm$  0.003 & 91.9  $\pm$  0.003 & 93.0 $\pm$  0.003 & 96.8  $\pm$  0.001 & 97.1  $\pm$  0.001\\
ARVGE & 92.4  $\pm$  0.004 & 92.6  $\pm$  0.004 & 92.4  $\pm$  0.003 & 93.0  $\pm$  0.003 & 96.5 $\pm$  0.001 & 96.8 $\pm$  0.001\\
\midrule
RWR-GAE & 92.9 $\pm$ 0.3 & 92.7 $\pm$ 0.5 & 92.1 $\pm$ 0.2 & 91.5 $\pm$ 0.08 & 96.2 $\pm$ 0.1 & 96.3 $\pm$ 0.09\\
RWR-VGAE & 92.6 $\pm$ 0.5 & 92.5 $\pm$ 0.7 & 92.3 $\pm$ 0.3 & 92.4 $\pm$ 0.1& 95.3 $\pm$ 0.1 & 95.2 $\pm$ 0.1\\
\bottomrule
\end{tabular}
\end{center}
\caption{Performance comparison of different models on the Link Prediction task across various datasets. We conduct each experiment 10 times and report the mean values with the standard deviation.}
\label{results:lp}
\end{table*}

\begin{table}[h]
\begin{center}
\begin{tabular}{lccccc}
\toprule
\bf Model & \bf Acc & \bf NMI & \bf F1 & \bf Precision & \bf ARI \\
\midrule
SC &  0.367 & 0.127 & 0.318 & 0.193 & 0.031\\
DW & 0.484 & 0.327 & 0.392 & 0.361 & 0.243\\
GAE & 0.596 & 0.429 & 0.595 & 0.596 & 0.347\\
VGAE & 0.609 & 0.436 & 0.609 & 0.609 & 0.346\\
ARGE & 0.640 & 0.449 & 0.619 & 0.646 & 0.352\\
ARVGE & 0.638 & 0.450 & 0.627 & 0.624 & 0.374\\
\midrule
RWR-GAE & 0.669	&\bf 0.481	& 0.618	& 0.629	&\bf 0.417\\
RWR-VGAE &\bf 0.685 & 0.455 &\bf 0.668 &\bf 0.685 &\bf 0.417\\
\bottomrule
\end{tabular}
\end{center}
\caption{Performance comparison of different models for the Clustering task on Cora.}
\label{results:nc:cora}
\end{table}

\begin{table}[h]
\begin{center}
\begin{tabular}{lccccc}
\toprule
\bf Model & \bf Acc & \bf NMI & \bf F1 & \bf Precision & \bf ARI \\
\midrule
SC & 0.239 &0.056 &0.299& 0.179 &0.010\\
DW & 0.337 &0.088 &0.270 &0.248 &0.092\\
GAE & 0.408 &0.176 &0.372 &0.418 &0.124\\
VGAE & 0.344 &0.156 &0.308 &0.349 &0.093\\
ARGE & 0.573 &0.350 &0.546 &0.573 &0.341\\
ARVGE & 0.544 &0.261 &0.529 &0.549 &0.245\\
\midrule
RWR-GAE &\bf 0.616	&\bf 0.354	&\bf 0.585	&\bf 0.605	&\bf 0.343\\
RWR-VGAE & 0.613	&0.338	&0.582	&0.595	&0.336\\
\bottomrule
\end{tabular}
\end{center}
\caption{Performance comparison of different models for the Clustering task on Citeseer.}
\label{results:nc:citeseer}
\end{table}

\begin{table}[h]
\begin{center}
\begin{tabular}{lccccc}
\toprule
\bf Model & \bf Acc & \bf NMI & \bf F1 & \bf Precision & \bf ARI \\
\midrule
GAE & 0.697&	0.33	&0.69&	0.72&	0.322\\
VGAE & 0.608	&0.219	&0.612	&0.613	&0.195\\
\midrule
RWR-GAE & 0.726	&\bf 0.355	&0.714	&0.729	&0.37\\
RWR-VGAE &\bf 0.736	&0.346	&\bf 0.725	&\bf 0.736	&\bf 0.381\\
\bottomrule
\end{tabular}
\end{center}
\caption{Performance comparison of different models for the Clustering task on PubMed.}
\label{results:nc:pubmed}
\end{table}

\section{Experimental Details}
\subsection{Datasets}
We report results on three benchmark graph datasets~\cite{sen2008collective}, Cora, Citeseer and pubMed. Each dataset is separated into a training, testing set and validation set. The validation set consists of 5\% citation edges for hyper-parameter tuning, the test set contains 10\% citation edges for reporting the final performance, and the rest are used for training. Table~\ref{stats} contains the training data statistics for each of the datatset.

\subsection{Tasks and Evaluation metric}
We evaluate the quality of the learned embeddings by analyzing the performance on two downstream tasks, Node clustering and Link Prediction.
\begin{itemize}
    \item \textbf{Node Clustering}, unsupervised clustering based on the node embeddings. After learning the embeddings, we do K-means clustering to get the final clusters. Following~\cite{xia2014robust}, we use five metrics to validate the clustering results: accuracy (Acc), normalized mutual information (NMI), precision, F-score (F1) and average rand index (ARI).
    \item \textbf{Link Prediction}, predict the edges and non-edges among the test set nodes. For doing such a prediction, we simply use Eq. \ref{reconstruct} to get the reconstructed graph from the node embeddings. Following \cite{vgae}, we report the AUC score (the area under a receiver operating characteristic curve) and average precision (AP) score. 
\end{itemize}

\subsection{Hyper-parameters}
We use the hyper-parameters provided by \cite{vgae} for the autoencoder related hyper-parameters of our model. For the hyper-parameters related to the Random Walk Regularization network, we set the number of walks to 50, window size to \{30, 20\} and walk length to \{30, 20\}. In our experiments, we find that the best performing model uses 50 walks with a window size and walk length of either 30 or 20 depending on the dataset and the kind of autoencoder.

\begin{figure*}[t]
\centering
\begin{subfigure}{.35\textwidth}
  \includegraphics[width=4.7cm, height=2.7cm]{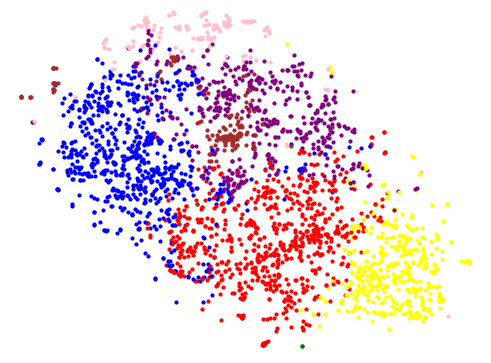}
\end{subfigure}%
\begin{subfigure}{.35\textwidth}
  \includegraphics[width=4.7cm, height=2.7cm]{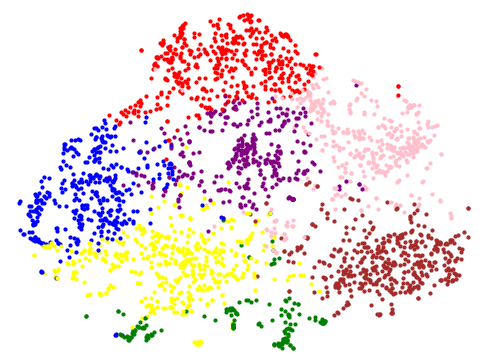}
\end{subfigure}%
\begin{subfigure}{.35\textwidth}
  \includegraphics[width=4.7cm, height=2.7cm]{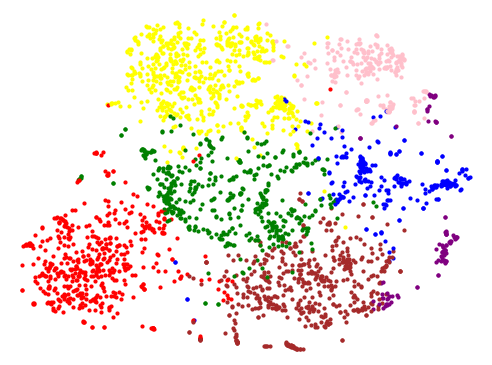}
\end{subfigure}%
\caption{Visualization using node embeddings generated by different hyperparameters using RWR-GAE on Cora. \textit{Left-to-right}: number of walks = walk length = window size = \{5, 20, 30\}, and accuracy = \{0.34, 0.56, 0.64\}}
\label{fig:vis:params}
\vspace{-1\baselineskip}
\end{figure*}

\section{Results}
We now present quantitative results of our model on the node clustering task. Tables \ref{results:nc:cora}, \ref{results:nc:citeseer} and \ref{results:nc:pubmed} contain the results for the datasets, cora, citeseer and pubmed respectively. We see that our proposed random walk regularization consistently outperforms all other baselines for all evaluation metrics. For the Cora dataset, we find that RWR-based methods improve the accuracy by 41.5\% when compared to DeepWalk and by 12.4\% when compared to Variational Graph Autoencoder. On the Citeseer dataset, we find that RWR-GAE beats the adversarially regularized autoencoder method by 7.5\% on accuracy and by 7.1\% on F1 score. For the PubMed dataset which has the largest number of nodes but the smallest number of clusters, we find that our method improves upon the GAE by 18.3\% on ARI and by 7.5\% on NMI.

Result for the link prediction task can be found in Table \ref{results:lp}. We find that our proposed method performs at par with the existing baselines. It is interesting to note that the random walk regularized autoencoder convincingly outperforms the DeepWalk method, indicating that the random walk is very well complemented by the autoencoder methods.

\section{Analysis}

\subsection{Graph Visualization \label{graph:dis}}
Figure \ref{fig:example} shows the quality of the embeddings using 2-d tsne~\cite{van2014accelerating} plots. The left plot is obtained by using the node embeddings learned by GAE and the right plot shows the graph embeddings learned by our RWR-GAE model. We observe that both the methods do a good job of identifying clusters based on the node embeddings. However, if we look closely at the embeddings generated by the GAE model, we find that the representation of the intra-cluster nodes are quite similar in nature but are not equally distant from the cluster centroid. Where as the embeddings generated by our RWR-GAE model have a more even spread within the cluster i.e the embeddings within a cluster are similar to each other. To measure this property, we define intra-cluster distance as the sum of euclidean distance of each node in the cluster to its centroid, averaged across all the clusters. We find that the embeddings generated by our model have less intra-cluster distance (0.64) compared to embeddings generated by GAE (0.99). We argue that this property is induced by the random walk regularization as the individual node embeddings need to predict the context nodes within a neighborhood, thus during training phase, the node embedding will prefer to converge to a representation such that it's informative of its context nodes. This results in improved embeddings for the clustering task, as a slight overlap among nodes of different clusters will now have relatively less impact on the clustering accuracy compared to the case when the intra-cluster embeddings are too close to each other and not evenly spread.


\subsection{Study using different hyper-parameters}
In this study, we try to understand how random walk helps in regularizing the embeddings. The left most embeddings in Figure \ref{fig:vis:params} are generated by our model with window size, number of walks and walk length set to 5. We observe that for this case, some of the clusters are subsumed into other clusters and thus achieve a very low clustering accuracy. We believe that this happens because of the extremely low values of walk length and window size. An intuitive explanation for this is that a low window size limits a nodes capability to look at enough nodes to decide its cluster candidacy. As we increase the window size, we observe that the clusters start to get more distinct from each other.

\subsection{Side-effects of random walk regularization}
We list down two critical observations about our model. First, from Table \ref{results:lp}, we observe that our proposed model has a considerably higher variance in scores. We attribute this behaviour to the introduced randomness while selecting nodes for random walk during the training phase. Second, from Algorithm \ref{alg:algorithm}, we see that the number of updates made to the encoder weights are considerably higher than the GAE model, as a result our proposed model converges to the best accuracy in fewer pass over the entire data. We consistently see the best results at around 100 epochs as opposed to 200 epochs for GAE.

\section{Conclusion}
We began by inspecting the graph embeddings learnt by an autoencoder model. 
We identified that this model doesn't enforce any restriction on the latent distribution and just uses a reconstruction loss for training which might result in sub-optimal embeddings. Further, we observed how this translated in terms of intra-cluster distances. We proposed a random walk based regularization technique for graph autoencoders which addressed both the shortcomings by adding a skipgram objective, which enforces the latent representations to capture network's local topology as well as provide additional training signal. We validated the effectiveness of our method by evaluating the performance of the learned embeddings on two different tasks, node clustering and link prediction for three standard datasets, \textit{cora}, \textit{citeseer} and \textit{pubmed}.


\bibliographystyle{named}
\bibliography{ijcai19}
\end{document}